\documentclass[sn-mathphys-num]{sn-jnl}

\usepackage{graphicx}%
\usepackage{multirow}%
\usepackage{amsmath,amssymb,amsfonts}%
\usepackage{amsthm}%
\usepackage{mathrsfs}%
\usepackage[title]{appendix}%
\usepackage{xcolor}%
\usepackage{textcomp}%
\usepackage{manyfoot}%
\usepackage{booktabs}%

\usepackage{listings}%

\theoremstyle{thmstyleone}%
\theoremstyle{thmstyletwo}%

\theoremstyle{thmstylethree}%

\begin{document}

\title[Exact formulation of Neural Networks]{Mathematical Programming Models for Exact and Interpretable Formulation of  Neural Networks}


\author*[1]{\fnm{Masoud} \sur{Ataei}}\email{masoud.ataei@utoronto.ca}

\author[2]{\fnm{Edrin} \sur{Hasaj}}\email{edrin.hasaj@mail.utoronto.ca}

\author[3]{\fnm{Jacob} \sur{Gipp}}\email{jacob.gipp@mail.utoronto.ca}

\author[4]{\fnm{Sepideh} \sur{Forouzi}}\email{sepideh.forouzi@colorado.edu}

\affil[1,2,3]{\normalsize\orgdiv{Department of Mathematical and Computational Sciences},\\ \orgname{University of Toronto}, \state{Ontario}, \country{Canada}}

\affil[4]{\normalsize\orgdiv{College of Engineering and Applied Science}, \orgname{University of Colorado Boulder}, \state{Colorado}, \country{U.S.A.}}



\abstract{This paper presents a unified mixed-integer programming framework for training sparse and interpretable neural networks. We develop exact formulations for both fully connected and convolutional architectures by modeling nonlinearities such as ReLU activations through binary variables and encoding structural sparsity via filter- and layer-level pruning constraints. The resulting models integrate parameter learning, architecture selection, and structural regularization within a single optimization problem, yielding globally optimal solutions with respect to a composite objective that balances prediction accuracy, weight sparsity, and architectural compactness. The mixed-integer programming formulation accommodates piecewise-linear operations, including max pooling and activation gating, and permits precise enforcement of logic-based or domain-specific constraints. By incorporating considerations of interpretability, sparsity, and verifiability directly into the training process, the proposed framework bridges a range of research areas including explainable artificial intelligence, symbolic reasoning, and formal verification.
 }




\maketitle

\section{Introduction}
Deep neural networks have achieved remarkable performance across a variety of tasks, yet their internal decision-making processes often remain opaque. In high-stakes domains such as healthcare and finance, this lack of transparency is a serious concern, as trust, accountability, and regulatory compliance demand interpretable models~\cite{rudin2019stop}. Relying on black-box models in critical applications can lead to harmful outcomes, and there is a growing consensus that one should instead design inherently interpretable networks from the start.

A rich body of research in explainable AI has explored methods to understand or constrain neural network behavior. Broadly, two paradigms exist: post-hoc explanation techniques versus intrinsically interpretable model design~\cite{lipton2018mythos}. Post-hoc methods (e.g., saliency maps~\cite{simonyan2013deep}, LIME~\cite{ribeiro2016should}, SHAP~\cite{lundberg2017unified}) attempt to explain a trained network's predictions without altering its structure. However, such explanations can be unreliable and do not guarantee understanding of the true underlying decision rules.

In contrast, intrinsically interpretable approaches bake transparency into the model itself. Rule extraction algorithms aim to distill a trained neural network into human-understandable if-then decision rules or decision trees~\cite{andrews1995survey}. Neuro-symbolic methods incorporate logical reasoning or structured knowledge into neural architectures~\cite{garcez2019neural}, allowing models to reason over interpretable structures. Sparsity-driven methods (e.g., \(\ell_1\)-regularization, pruning) simplify networks by forcing many weights or neurons to zero, highlighting only the most relevant features~\cite{han2015deep,lecun1990optimal}.

In parallel, mathematical programming, particularly mixed-integer linear programming (MILP), has emerged as a tool for encoding neural network behavior, enabling verification and certification tasks~\cite{tjeng2017evaluating,serra2020bounding,anderson2020strong, nair2020solving, dua2010mixed, thorbjarnarson2023optimal,schwan2023stability}. MILP formulations have been used to verify adversarial robustness~\cite{fischetti2018deep}, enforce fairness~\cite{perez2022fair}, and compress networks optimally~\cite{el2020oamip}. However, most existing formulations target specific objectives (e.g., robustness verification or post-hoc pruning) and do not simultaneously support interpretability constraints during training.

This work presents a unified MILP framework for designing interpretable neural architectures. We encode feed-forward and convolutional layers using exact linear constraints for ReLU activations and affine transformations, while introducing binary variables to model neuron and layer selection. Our formulation supports structured pruning, sparsity promotion, and exact verifiability. By integrating training objectives and structural regularization into a single MILP, we achieve interpretable architectures optimized under verifiable conditions.

The remainder of this paper is organized as follows. Section~\ref{Sec:MIP_Dense} presents a mixed-integer linear programming formulation for dense, fully connected feed-forward neural networks, with explicit modeling of nonlinear activation functions such as ReLU. This section also discusses considerations related to sparsity and interpretability of the resulting models. In Section~\ref{Sec:MILP_CNN}, the framework is extended to convolutional neural networks through the introduction of constraints that encode convolutional operations, max-pooling layers, and activation gating. Section~\ref{Sec:Conclusion} concludes the paper with a discussion of prospective applications and possible directions for extending the proposed MILP-based approach.

\section{MILP Model of Dense Neural Networks}
\label{Sec:MIP_Dense}

We present an MILP formulation that captures the exact behavior of a fully connected feed-forward neural network, enabling simultaneous training, architecture selection, and structural sparsity enforcement. The model encodes affine transformations, piecewise-linear ReLU activations, and layer-level pruning through a unified system of linear constraints and binary variables. This allows the network to be optimized globally with respect to a composite objective balancing predictive accuracy, parameter sparsity, and architectural compactness.

To begin, we define the core notation and structural elements of the network. Let the training dataset be denoted by
\[
\mathcal{D} = \{(\mathbf{x}_i, \mathbf{t}_i)\}_{i=1}^n \subset \mathbb{R}^{n_0} \times \mathbb{R}^{n_{L+1}},
\]
where \( \mathbf{x}_i \in \mathbb{R}^{n_0} \) is the input and \( \mathbf{t}_i \in \mathbb{R}^{n_{L+1}} \) is the target output of the \( i \)-th sample. The network comprises \( L \in \mathbb{Z}_{>0} \) hidden layers and one output layer, indexed by \( l \in [L+1] := \{1, 2, \ldots, L+1\} \).

Each layer \( l \) contains \( n_l \) neurons and is parameterized by a weight matrix \( \mathbf{W}^{(l)} \in \mathbb{R}^{n_l \times n_{l-1}} \) and bias vector \( \mathbf{b}^{(l)} \in \mathbb{R}^{n_l} \). For a given input \( \mathbf{x}_i \), the network computes intermediate pre-activations \( \mathbf{z}_i^{(l)} \in \mathbb{R}^{n_l} \) and activations \( \mathbf{a}_i^{(l)} \in \mathbb{R}^{n_l} \) at each layer. The ReLU activation function is applied to hidden layers and is defined component-wise by
\[
\operatorname{ReLU}(z) := \max(0, z).
\]
To exactly model this nonlinearity within the MILP framework, we introduce binary variables that encode the activation regime of each neuron per sample, yielding a precise piecewise-linear characterization of the network's forward computation. The full optimization model is then given as follows:

\begin{subequations} \label{eq:mip_vector}
	\begin{align}
		\min \quad 
		& \sum_{i=1}^n \left\| \mathbf{a}_i^{(L+1)} - \mathbf{t}_i \right\|_2^2 
		+ \alpha \lambda \sum_{l=1}^{L+1} \sum_{j=1}^{n_l} \sum_{k=1}^{n_{l-1}} u_{j,k}^{(l)} \nonumber \\
		& + \tfrac{1}{2} \alpha (1 - \lambda) \sum_{l=1}^{L+1} \left\| \mathbf{W}^{(l)} \right\|_F^2 
		+ \beta \sum_{l=1}^L \gamma_l \label{eq:objective_function} \\		
		\text{s.t.} \quad 
		& \mathbf{a}_i^{(0)} = \mathbf{x}_i, && \forall i \in [n] \label{eq:input_assignment} \\		
		& \mathbf{z}_i^{(l)} = \mathbf{W}^{(l)} \mathbf{a}_i^{(l-1)} + \mathbf{b}^{(l)}, 
		&& \forall i \in [n],\ \forall l \in [L] \label{eq:affine_map} \\		
		& \mathbf{a}_i^{(L+1)} = \mathbf{W}^{(L+1)} \mathbf{a}_i^{(L)} + \mathbf{b}^{(L+1)}, 
		&& \forall i \in [n] \label{eq:output_map} \\		
		& a_{i,j}^{(l)} \geq 0, 
		&& \forall i \in [n],\ \forall l \in [L],\ \forall j \in [n_l] \label{eq:relu_lower} \\		
		& a_{i,j}^{(l)} \geq z_{i,j}^{(l)}, 
		&& \forall i \in [n],\ \forall l \in [L],\ \forall j \in [n_l] \label{eq:relu_identity_lb} \\		
		& a_{i,j}^{(l)} \leq z_{i,j}^{(l)} - z_{\min}^{(l)}(1 - \delta_{i,j}^{(l)}), 
		&& \forall i \in [n],\ \forall l \in [L],\ \forall j \in [n_l] \label{eq:relu_identity_ub} \\		
		& a_{i,j}^{(l)} \leq z_{\max}^{(l)} \delta_{i,j}^{(l)}, 
		&& \forall i \in [n],\ \forall l \in [L],\ \forall j \in [n_l] \label{eq:relu_activation_bound} \\		
		& 0 \leq a_{i,j}^{(l)},\ z_{i,j}^{(l)} \leq M \gamma_l, 
		&& \forall i \in [n],\ \forall l \in [L],\ \forall j \in [n_l] \label{eq:pruning_activation} \\		
		& u_{j,k}^{(l)} \geq W_{j,k}^{(l)},\quad u_{j,k}^{(l)} \geq -W_{j,k}^{(l)}, 
		&& \forall l \in [L+1],\ \forall j \in [n_l],\ \forall k \in [n_{l-1}] \label{eq:l1_linearization} \\		
		& -M \gamma_l \leq W_{j,k}^{(l)} \leq M \gamma_l, 
		&& \forall l \in [L],\ \forall j \in [n_l],\ \forall k \in [n_{l-1}] \label{eq:prune_weights} \\		
		& -M \gamma_l \leq b_j^{(l)} \leq M \gamma_l, 
		&& \forall l \in [L],\ \forall j \in [n_l] \label{eq:prune_biases} \\		
		& \gamma_{l+1} \leq \gamma_l, 
		&& \forall l \in [L-1] \label{eq:layer_ordering} \\		
		& \gamma_1 = 1 \label{eq:root_layer_active} \\		
		& \sum_{k=1}^{n_{l-1}} W_{j,k}^{(l)} \geq \sum_{k=1}^{n_{l-1}} W_{j+1,k}^{(l)}, 
		&& \forall l \in [L],\ \forall j \in [n_l - 1] \label{eq:symmetry_breaking} \\		
		& \delta_{i,j}^{(l)} \in \{0,1\}, 
		&& \forall i \in [n],\ \forall l \in [L],\ \forall j \in [n_l] \label{eq:relu_binary} \\		
		& \gamma_l \in \{0,1\}, 
		&& \forall l \in [L] \label{eq:layer_binary} \\		
		& u_{j,k}^{(l)} \geq 0,\quad 
		&& \forall l \in [L+1],\ \forall j \in [n_l],\ \forall k \in [n_{l-1}] \label{eq:aux_nonneg}
	\end{align}
\end{subequations}

The forward computation in a dense neural network is initialized by assigning each input vector \( \mathbf{x}_i \) to the activation of the input layer. This anchoring step is encoded in constraint~\eqref{eq:input_assignment} as follows:
\[
\mathbf{a}_i^{(0)} = \mathbf{x}_i, \quad \forall i \in [n].
\]
It establishes the input activations as the foundation for all subsequent propagation through the network layers.

For every hidden layer \( l \in [L] \), the pre-activation vector \( \mathbf{z}_i^{(l)} \in \mathbb{R}^{n_l} \) is computed via an affine transformation of the previous layer's activations, as expressed by constraint~\eqref{eq:affine_map} as follows:
\[
\mathbf{z}_i^{(l)} = \mathbf{W}^{(l)} \mathbf{a}_i^{(l-1)} + \mathbf{b}^{(l)}, \quad \forall i \in [n].
\]
Here, \( \mathbf{W}^{(l)} \) and \( \mathbf{b}^{(l)} \) denote the learnable weights and biases associated with layer \( l \). This operation linearly aggregates the incoming signals to produce the pre-activation inputs for each neuron in the layer.

The output of the final layer \( L+1 \), which is typically linear to support both regression and classification tasks, is similarly computed by constraint~\eqref{eq:output_map} through
\[
\mathbf{a}_i^{(L+1)} = \mathbf{W}^{(L+1)} \mathbf{a}_i^{(L)} + \mathbf{b}^{(L+1)}, \quad \forall i \in [n].
\]
Together, constraints~\eqref{eq:input_assignment} through \eqref{eq:output_map} define the deterministic flow of activations through the network, forming the structural backbone of the MILP model.

Nonlinear behavior is introduced via the ReLU function, \( \operatorname{ReLU}(z) = \max(0, z) \), encoded exactly using binary variables \( \delta_{i,j}^{(l)} \in \{0,1\} \) that indicate whether a hidden unit is active. Constraints~\eqref{eq:relu_lower}-\eqref{eq:relu_activation_bound} together implement this piecewise-linear mapping.

Constraint~\eqref{eq:relu_lower} enforces non-negativity of the activation, while \eqref{eq:relu_identity_lb} ensures that it lower-bounds the pre-activation. Constraint~\eqref{eq:relu_identity_ub} enables identity behavior only when \( \delta_{i,j}^{(l)} = 1 \), and \eqref{eq:relu_activation_bound} forces the activation to zero when \( \delta_{i,j}^{(l)} = 0 \). Collectively, these constraints yield the exact relation
\[
a_{i,j}^{(l)} = \max(0, z_{i,j}^{(l)}), \quad \forall i \in [n],\ l \in [L],\ j \in [n_l],
\]
with \( \delta_{i,j}^{(l)} \) controlling the activation regime. The constants \( z_{\min}^{(l)} < 0 \) and \( z_{\max}^{(l)} > 0 \) are precomputed bounds on pre-activations, critical for maintaining tight big-\( M \) constraints and mitigating relaxation error.

The output layer \( L+1 \), governed by constraint~\eqref{eq:output_map}, omits nonlinearity; i.e.,
\[
\mathbf{a}_i^{(L+1)} = \mathbf{W}^{(L+1)} \mathbf{a}_i^{(L)} + \mathbf{b}^{(L+1)}, \quad \forall i \in [n],
\]
supporting both regression and classification, with softmax or thresholding applied externally if needed.

Thus, constraints~\eqref{eq:input_assignment}-\eqref{eq:output_map} collectively define the exact forward propagation path within the MILP framework, integrating linear transformations and ReLU activations into a tractable, symbolic optimization model.

To promote connection-level sparsity, the formulation introduces auxiliary continuous variables \( u_{j,k}^{(l)} \geq 0 \), constrained by
\[
u_{j,k}^{(l)} \geq W_{j,k}^{(l)}, \quad u_{j,k}^{(l)} \geq -W_{j,k}^{(l)},
\]
as specified in constraint~\eqref{eq:l1_linearization}. At optimality, each auxiliary variable satisfies \( u_{j,k}^{(l)} = |W_{j,k}^{(l)}| \), enabling an exact linearization of the \( \ell_1 \)-norm. These variables appear in the regularization term of the objective~\eqref{eq:objective_function}, scaled by \( \alpha \lambda \), thereby penalizing total connection strength and encouraging weight-level sparsity. Constraint~\eqref{eq:aux_nonneg} enforces their non-negativity, ensuring correctness of the linearization. Since no explicit upper bounds are imposed, tightness is driven purely by minimization, yielding sparse and interpretable parameterizations.

To enable architectural sparsity, binary variables \( \gamma_l \in \{0,1\} \) govern layer-level pruning. When \( \gamma_l = 0 \), all weights, biases, and activations of layer \( l \) are nulled via constraints~\eqref{eq:prune_weights}, \eqref{eq:prune_biases}, and \eqref{eq:pruning_activation}, effectively removing the layer. The ReLU system (constraints~\eqref{eq:relu_lower}-\eqref{eq:relu_activation_bound}) naturally resolves to zero when the input vanishes, maintaining functional consistency. A sparsity-inducing penalty term \( \beta \sum_{l=1}^L \gamma_l \) in the objective discourages unnecessary layers, complementing the \( \ell_1 \)-based weight regularization and promoting compact architectures.

To preserve coherent topological structure, the ordering constraint~\eqref{eq:layer_ordering} ensures that if layer \( l \) is pruned, then all deeper layers \( l' > l \) must also be pruned. Constraint~\eqref{eq:root_layer_active} enforces \( \gamma_1 = 1 \), guaranteeing the presence of at least one hidden layer and ruling out degenerate architectures. Together, these constraints allow the MILP to determine network depth adaptively, yielding architectures that are minimal, valid, and interpretable. This binary structural regularization distinguishes the formulation from continuous relaxation-based approaches, enabling the simultaneous discovery of sparse connections and shallow topologies within a globally optimal training framework.

To reduce redundant exploration of isomorphic solutions, we impose the symmetry-breaking constraint~\eqref{eq:symmetry_breaking}, which lexicographically orders the rows of each hidden layer's weight matrix. For each layer \( l \in [L] \), the row sums of \( \mathbf{W}^{(l)} \) are required to be non-increasing; i.e.,
\[
\sum_{k=1}^{n_{l-1}} W_{j,k}^{(l)} \geq \sum_{k=1}^{n_{l-1}} W_{j+1,k}^{(l)}, \quad \forall j \in [n_l - 1].
\]
This constraint eliminates symmetric representations arising from permutations of hidden units, which are functionally equivalent but inflate the MILP search space. Enforcing a canonical ordering restricts the solver to a reduced, non-redundant subset of feasible architectures, thereby improving convergence without affecting optimality.

The objective function~\eqref{eq:objective_function} combines four components that jointly govern predictive accuracy, parameter sparsity, and architectural simplicity.

The first term,
\[
\sum_{i=1}^n \left\| \mathbf{a}_i^{(L+1)} - \mathbf{t}_i \right\|_2^2,
\]
represents the squared error loss between predicted outputs \( \mathbf{a}_i^{(L+1)} \) and ground-truth labels \( \mathbf{t}_i \). This term enforces empirical fidelity, serving as the standard loss for regression and a quadratic surrogate to cross-entropy in classification when one-hot targets are used.

The second and third terms implement a convex regularization scheme. The linearized \( \ell_1 \)-norm,
\[
\alpha \lambda \sum_{l=1}^{L+1} \sum_{j=1}^{n_l} \sum_{k=1}^{n_{l-1}} u_{j,k}^{(l)},
\]
penalizes connection density and is enforced via constraint~\eqref{eq:l1_linearization}, which guarantees \( u_{j,k}^{(l)} = |W_{j,k}^{(l)}| \) at optimality. This promotes sparsity at the edge level, enhancing interpretability and compression.

Complementing this, the Frobenius norm penalty,
\[
\frac{1}{2} \alpha (1 - \lambda) \sum_{l=1}^{L+1} \left\| \mathbf{W}^{(l)} \right\|_F^2,
\]
discourages large weight magnitudes and improves generalization. The hyperparameters \( \alpha > 0 \) and \( \lambda \in [0,1] \) jointly control the strength and balance of sparsity (\( \ell_1 \)) versus shrinkage (\( \ell_2 \)).

The final term,
\[
\beta \sum_{l=1}^L \gamma_l,
\]
penalizes architectural complexity via binary layer indicators \( \gamma_l \in \{0,1\} \) (see ~\eqref{eq:layer_binary}). When \( \gamma_l = 0 \), layer \( l \) is pruned via constraints~\eqref{eq:prune_weights}-\eqref{eq:pruning_activation}. The parameter \( \beta > 0 \) governs the trade-off between depth and parsimony.

Altogether, this Lagrangian-style objective balances data fit, sparsity, and structural simplicity. By tuning \( \alpha \), \( \lambda \), and \( \beta \), the model interpolates between expressive and highly interpretable configurations.

The proposed formulation supports multiple forms of sparsity. Connection-level sparsity is induced directly by the \( \ell_1 \)-regularization term. Neuron-level sparsity arises implicitly: if all incoming weights to a neuron vanish, its activation becomes identically zero. Layer-level sparsity is enforced explicitly through the binary switches \( \gamma_l \), which disable all computations associated with a layer.

Despite the presence of binary variables \( \delta_{i,j}^{(l)} \) and \( \gamma_l \), the model remains tractable for moderate-scale problems due to its modular, linear structure. Domain-specific constraints, such as fairness, monotonicity, or logical conditions, can be seamlessly incorporated as additional linear constraints, enabling verifiable and interpretable neural network learning.

By minimizing the composite objective~\eqref{eq:objective_function}, the MILP formulation jointly addresses training and architecture selection. The binary ReLU indicators \( \delta_{i,j}^{(l)} \) enable per-sample activation modeling, allowing the solver to explore the exponential space of possible activation patterns. This global perspective contrasts with local, gradient-based training, permitting structural choices such as neuron or layer deactivation via zeroed weights or inactive switches \( \gamma_l \).

A key strength of the formulation lies in its theoretical precision. Each feasible MILP solution corresponds to a fully specified neural network, with exact weights, activations, and topology. If optimal, it minimizes empirical error subject to explicit sparsity and architecture constraints, offering guarantees that are generally unavailable in heuristic deep learning pipelines.

Naturally, this expressiveness incurs computational cost. Solving MILPs is NP-hard, and the number of binary variables grows with both the number of samples and hidden units. Nevertheless, the formulation remains solvable for moderate problem sizes. State-of-the-art solvers (e.g., Gurobi, CPLEX) exploit relaxations, cutting planes, and symmetry-breaking constraints such as~\eqref{eq:symmetry_breaking} to improve efficiency. Regularization parameters \( \alpha \) and \( \beta \) further enhance tractability by promoting sparsity and reducing the effective model complexity.

The linear structure of the model offers significant extensibility. Properties like fairness, logical consistency, or monotonicity can be enforced directly as linear constraints without sacrificing solvability or interpretability. This makes the MILP framework not just a training algorithm, but a certifiable platform for trustworthy and transparent neural network design.

It is also worth noting that the formulation uses \( O(n \sum_l n_l) \) continuous variables for activations, \( O(\sum_l n_l n_{l-1}) \) for weights and biases, and \( O(n \sum_l n_l) \) binary variables for ReLU states, along with \( L \) binary variables for layer pruning and auxiliary variables \( u_{j,k}^{(l)} \). While this dimensionality limits scalability to large architectures, the model remains highly effective for moderate-sized networks, yielding globally optimal, interpretable, and verifiable neural representations.

\section{MILP Model of Convolutional Neural Networks}
\label{Sec:MILP_CNN}

We now extend the MLIP framework to model convolutional neural networks (CNN), incorporating convolutional operations, ReLU activations, and channel-level pruning. The formulation remains fully deterministic and encodes nonlinear behavior via exact piecewise-linear constraints. Convolutional structure is modeled using tensor-valued variables, while pruning and sparsity are enforced through binary and auxiliary variables, respectively.

Let the training dataset be denoted by $$ \mathcal{D} = \{ (\mathcal{X}_i, \mathbf{t}_i) \}_{i=1}^n, $$ where each input tensor \( \mathcal{X}_i \in \mathbb{R}^{C_0 \times H_0 \times W_0} \) represents a multi-channel image and \( \mathbf{t}_i \in \mathbb{R}^{n_{L+1}} \) denotes its associated label vector. The network consists of \( L \in \mathbb{Z}_{>0} \) convolutional layers. Each layer \( l \in [L] \) is parameterized by a kernel tensor \( \mathcal{W}^{(l)} \in \mathbb{R}^{C_l \times C_{l-1} \times K_H^{(l)} \times K_W^{(l)} } \) and bias vector \( \mathbf{b}^{(l)} \in \mathbb{R}^{C_l} \), yielding pre-activation and activation tensors \( \mathcal{Z}_i^{(l)} \) and \( \mathcal{A}_i^{(l)} \), respectively, with shape \( \mathbb{R}^{C_l \times H_l \times W_l} \).

ReLU activations are modeled by binary variables \( \delta_{i,c,h,w}^{(l)} \in \{0,1\} \), which encode spatially localized activation states for each channel. Output-channel pruning is supported via binary switches \( \gamma_c^{(l)} \in \{0,1\} \), which deactivate all spatial locations associated with channel \( c \) in layer \( l \). To enable \( \ell_1 \)-based regularization of convolutional weights, we introduce nonnegative auxiliary variables \( u_{c,c',u,v}^{(l)} \geq 0 \), which linearize the absolute value of each kernel entry.

The resulting MILP formulation jointly models spatial convolution, activation switching, sparsity, and structural pruning in a unified optimization framework, which is given as follows:

\begin{subequations} \label{eq:mip_cnn}
	\begin{align}
		\min \quad & \sum_{i=1}^n \left\| \mathbf{a}_i^{(L+1)} - \mathbf{t}_i \right\|_2^2 + \alpha \lambda \sum_{l=1}^{L+1} \sum_{c,c',u,v} u_{c,c',u,v}^{(l)}  \nonumber \\
		&+ \tfrac{\alpha}{2}(1 - \lambda) \sum_{l=1}^{L+1} \left\| \mathcal{W}^{(l)} \right\|_F^2 + \beta \sum_{l=1}^{L} \sum_{c=1}^{C_l} \gamma_c^{(l)} \label{eq:mip_cnn_obj} \\
		\text{s.t.} \quad & \mathcal{A}_i^{(0)} = \mathcal{X}_i, && \forall i \in [n] \label{eq:mip_cnn_input} \\
		& z_{i,c,h,w}^{(l)} = \sum_{c'} \sum_{u,v} \mathcal{W}_{c,c',u,v}^{(l)} a_{i,c',h+u,w+v}^{(l-1)} + b_c^{(l)}, && \forall i \in [n],\ l \in [L],\ c \in [C_l],\ (h,w) \in \mathcal{I}_l \label{eq:mip_cnn_conv} \\
		& a_{i,c,h,w}^{(l)} \geq 0, && \forall i \in [n],\ l \in [L],\ c \in [C_l],\ (h,w) \in \mathcal{I}_l \label{eq:mip_cnn_relu1} \\
		& a_{i,c,h,w}^{(l)} \geq z_{i,c,h,w}^{(l)}, && \forall i \in [n],\ l \in [L],\ c \in [C_l],\ (h,w) \in \mathcal{I}_l \label{eq:mip_cnn_relu2} \\
		& a_{i,c,h,w}^{(l)} \leq z_{i,c,h,w}^{(l)} - z_{\min}^{(l)} (1 - \delta_{i,c,h,w}^{(l)}), && \forall i \in [n],\ l \in [L],\ c \in [C_l],\ (h,w) \in \mathcal{I}_l \label{eq:mip_cnn_relu3} \\
		& a_{i,c,h,w}^{(l)} \leq z_{\max}^{(l)} \delta_{i,c,h,w}^{(l)}, && \forall i \in [n],\ l \in [L],\ c \in [C_l],\ (h,w) \in \mathcal{I}_l \label{eq:mip_cnn_relu4} \\
		& \sum_{(h,w) \in \Omega_{c,h',w'}^{(l)}} \zeta_{i,c,h,w}^{(l)} = 1, && \forall i \in [n],\ l \in \mathcal{L}_{\mathrm{pool}},\ c \in [C_l],\ (h',w') \in \mathcal{I}_l^{\mathrm{pool}} \label{eq:mip_cnn_maxpool1} \\
		& p_{i,c,h',w'}^{(l)} \geq a_{i,c,h,w}^{(l)}, && \forall i,\ l \in \mathcal{L}_{\mathrm{pool}},\ c,\ (h',w') \in \mathcal{I}_l^{\mathrm{pool}},\ (h,w) \in \Omega_{c,h',w'}^{(l)} \label{eq:mip_cnn_maxpool2} \\
		& p_{i,c,h',w'}^{(l)} \leq a_{i,c,h,w}^{(l)} + M (1 - \zeta_{i,c,h,w}^{(l)}), && \forall i,\ l \in \mathcal{L}_{\mathrm{pool}},\ c,\ (h',w') \in \mathcal{I}_l^{\mathrm{pool}},\ (h,w) \in \Omega_{c,h',w'}^{(l)} \label{eq:mip_cnn_maxpool3} \\
		& a_{i,f}^{(L)} = a_{i,c,h,w}^{(L)}, && \forall i,\ f = c \cdot H_L \cdot W_L + h \cdot W_L + w \label{eq:mip_cnn_flatten} \\
		& \mathbf{a}_i^{(L+1)} = \mathbf{W}^{(L+1)} \mathbf{a}_i^{(L)} + \mathbf{b}^{(L+1)}, && \forall i \in [n] \label{eq:mip_cnn_output} \\
		& u_{c,c',u,v}^{(l)} \geq \mathcal{W}_{c,c',u,v}^{(l)},\quad u_{c,c',u,v}^{(l)} \geq -\mathcal{W}_{c,c',u,v}^{(l)}, && \forall l \in [L+1],\ c,c',u,v \label{eq:mip_cnn_l1} \\
		& |\mathcal{W}_{c,c',u,v}^{(l)}| \leq M \gamma_c^{(l)},\quad |b_c^{(l)}| \leq M \gamma_c^{(l)}, && \forall l \in [L],\ c,c',u,v \label{eq:mip_cnn_prune1} \\
		& 0 \leq z_{i,c,h,w}^{(l)},\ a_{i,c,h,w}^{(l)} \leq M \gamma_c^{(l)}, && \forall i \in [n],\ l \in [L],\ c \in [C_l],\ (h,w) \in \mathcal{I}_l \label{eq:mip_cnn_prune2} \\
		& \sum_{c',u,v} u_{c,c',u,v}^{(l)} \geq \sum_{c',u,v} u_{c+1,c',u,v}^{(l)}, && \forall l \in [L],\ c \in [C_l - 1] \label{eq:mip_cnn_symmetry} \\
		& \mathcal{W}_{c,c',u,v}^{(l)} \in [-M, M],\quad b_c^{(l)} \in [-M, M], && \forall l,\ c,\ c',\ u,\ v \label{eq:mip_cnn_box1} \\
		& z_{i,c,h,w}^{(l)},\ a_{i,c,h,w}^{(l)},\ p_{i,c,h',w'}^{(l)} \in [0, M], && \forall i,\ c,\ h,\ w,\ l \in [L] \label{eq:mip_cnn_box2} \\
		& u_{c,c',u,v}^{(l)} \geq 0, && \forall l,\ c,\ c',\ u,\ v \label{eq:mip_cnn_aux_nonneg} \\
		& \delta_{i,c,h,w}^{(l)} \in \{0,1\}, && \forall i,\ l \in [L],\ c \in [C_l],\ (h,w) \in \mathcal{I}_l \label{eq:mip_cnn_binary_relu} \\
		& \xi_{i,c,h,w}^{(l)} \in \{0,1\}, && \forall i,\ l \in [L],\ c \in [C_l],\ (h,w) \in \mathcal{I}_l \label{eq:mip_cnn_binary_ksi} \\
		& \gamma_c^{(l)} \in \{0,1\}, && \forall l \in [L],\ c \in [C_l] \label{eq:mip_cnn_binary_layer}
	\end{align}
\end{subequations}

The initialization of the convolutional neural network is enforced by constraint~\eqref{eq:mip_cnn_input}, which sets the zeroth-layer activation tensor equal to the input: \( \mathcal{A}_i^{(0)} = \mathcal{X}_i \), for all data samples \( i \in [n] \). This constraint anchors the network at the input layer and ensures that subsequent transformations operate on a well-defined, data-dependent tensor. The notation \( \mathcal{A}_i^{(0)} \in \mathbb{R}^{C_0 \times H_0 \times W_0} \) denotes the activation of the zeroth layer for instance \( i \), preserving the full spatial and channel-wise structure of the input image throughout the early stages of the convolutional hierarchy.

The convolutional operation in the forward pass is encoded by constraint~\eqref{eq:mip_cnn_conv}, which defines the pre-activation value \( z_{i,c,h,w}^{(l)} \) for each data instance \( i \in [n] \), convolutional layer \( l \in [L] \), output channel \( c \in [C_l] \), and spatial location \( (h,w) \in \mathcal{I}_l \). The computation aggregates contributions from a local receptive field in the previous layer, yielding
\[
z_{i,c,h,w}^{(l)} = \sum_{c'} \sum_{u,v} \mathcal{W}_{c,c',u,v}^{(l)} a_{i,c',h+u,w+v}^{(l-1)} + b_c^{(l)}.
\]
Here, \( \mathcal{W}_{c,c',u,v}^{(l)} \in \mathbb{R} \) denotes the kernel weight applied between input channel \( c' \in [C_{l-1}] \) and output channel \( c \), with spatial offset \( (u,v) \in [K_H^{(l)}] \times [K_W^{(l)}] \). The scalar bias \( b_c^{(l)} \in \mathbb{R} \) is shared across spatial positions for each output channel. The index set \( \mathcal{I}_l \subseteq \mathbb{Z}_{\geq 0} \times \mathbb{Z}_{\geq 0} \) encodes valid spatial positions of the output feature map, determined by the convolution stride, padding, and kernel size. By explicitly unrolling the convolution into a linear affine transformation, the formulation preserves compatibility with the MILP framework while capturing the spatial and channel-wise locality inherent to convolutional architectures.

Following the computation of pre-activations, nonlinear transformations are introduced via the ReLU activation function, defined as \( \max(0, z) \). The MILP formulation encodes this nonlinearity exactly using constraints~\eqref{eq:mip_cnn_relu1} through~\eqref{eq:mip_cnn_relu4}, which collectively govern the post-activation values \( a_{i,c,h,w}^{(l)} \). Central to this encoding is the binary variable \( \delta_{i,c,h,w}^{(l)} \in \{0,1\} \), which determines whether the ReLU is in its active (identity) or inactive (zero) regime at a given unit.

Constraint~\eqref{eq:mip_cnn_relu1} enforces the nonnegativity of all activations, while constraint~\eqref{eq:mip_cnn_relu2} ensures that each activation lower-bounds its corresponding pre-activation. Constraint~\eqref{eq:mip_cnn_relu3} imposes an upper bound of the form \( z_{i,c,h,w}^{(l)} - z_{\min}^{(l)}(1 - \delta_{i,c,h,w}^{(l)}) \), which becomes tight when \( \delta_{i,c,h,w}^{(l)} = 1 \), forcing the activation to match the pre-activation. When \( \delta_{i,c,h,w}^{(l)} = 0 \), the term relaxes to a non-binding constraint, due to the assumed bound \( z_{\min}^{(l)} < 0 \). Complementarily, constraint~\eqref{eq:mip_cnn_relu4} enforces \( a_{i,c,h,w}^{(l)} \leq z_{\max}^{(l)} \delta_{i,c,h,w}^{(l)} \), ensuring that the activation collapses to zero when the ReLU is inactive. This last constraint is trivially satisfied when \( \delta_{i,c,h,w}^{(l)} = 1 \), since the remaining constraints already enforce \( a = z \).

Taken together, these four constraints yield an exact piecewise-linear representation of the ReLU function within the MILP framework. The binary variable \( \delta_{i,c,h,w}^{(l)} \) precisely encodes the switch between the two activation regimes, ensuring that the equality \( a_{i,c,h,w}^{(l)} = \max(0, z_{i,c,h,w}^{(l)}) \) is satisfied for all layers \( l \in [L] \), channels \( c \in [C_l] \), and spatial positions \( (h,w) \in \mathcal{I}_l \). The constants \( z_{\min}^{(l)} < 0 \) and \( z_{\max}^{(l)} > 0 \) are precomputed per layer to reflect conservative bounds on the pre-activations and serve to tighten the big-\( M \) terms. Their careful selection is essential for improving numerical stability and reducing relaxation slack within the MILP solver.

Max-pooling operations are modeled explicitly via a set of linear constraints that identify and extract the maximal value within a local receptive field. For each data instance \( i \in [n] \), layer \( l \in \mathcal{L}_{\mathrm{pool}} \), channel \( c \in [C_l] \), and pooled spatial location \( (h', w') \in \mathcal{I}_l^{\mathrm{pool}} \), the MILP formulation introduces binary selection variables \( \zeta_{i,c,h,w}^{(l)} \in \{0,1\} \), which determine the active site within the pooling window \( \Omega_{c,h',w'}^{(l)} \subseteq \mathcal{I}_l \) from which the maximum is drawn.

Constraint~\eqref{eq:mip_cnn_maxpool1} enforces that exactly one such location is selected as follows:
\[
\sum_{(h,w) \in \Omega_{c,h',w'}^{(l)}} \zeta_{i,c,h,w}^{(l)} = 1,
\]
ensuring that the pooled output is uniquely determined. Constraint~\eqref{eq:mip_cnn_maxpool2} guarantees that the pooled activation \( p_{i,c,h',w'}^{(l)} \) dominates all candidates in the pooling region, by enforcing
\[
p_{i,c,h',w'}^{(l)} \geq a_{i,c,h,w}^{(l)} \quad \forall (h,w) \in \Omega_{c,h',w'}^{(l)}.
\]
Constraint~\eqref{eq:mip_cnn_maxpool3} completes the encoding by bounding the pooled output from above using a big-\( M \) expression as follows:
\[
p_{i,c,h',w'}^{(l)} \leq a_{i,c,h,w}^{(l)} + M(1 - \zeta_{i,c,h,w}^{(l)}),
\]
which becomes binding only when \( \zeta_{i,c,h,w}^{(l)} = 1 \), i.e., when the activation at \( (h,w) \) is selected as the maximal value. For all other candidates, the constraint relaxes via the large constant \( M \), which must be chosen conservatively to dominate the feasible range of activations.

Together, these three constraints exactly encode the max-pooling operation within a linear-integer framework. While the use of binary selection variables \( \zeta \) introduces a combinatorial component, this representation preserves the discrete behavior of max pooling and ensures functional equivalence with its neural counterpart. The ability to model such localized nonconvex operations using linear constraints is essential to maintaining the expressive capacity of the convolutional architecture in the MILP setting.

The final stage of the network computation is defined by constraint~\eqref{eq:mip_cnn_output}, which models the fully connected output layer. At this point, the activation tensor \( \mathcal{A}_i^{(L)} \in \mathbb{R}^{C_L \times H_L \times W_L} \), produced by the last convolutional or pooling layer, is flattened into a vector \( \mathbf{a}_i^{(L)} \in \mathbb{R}^{C_L H_L W_L} \) as specified by constraint~\eqref{eq:mip_cnn_flatten}. This flattening is performed using a channel-major indexing convention
\[
a_{i,f}^{(L)} = a_{i,c,h,w}^{(L)}, \qquad \text{where } f = c \cdot H_L \cdot W_L + h \cdot W_L + w,
\]
ensuring that each triplet \( (c, h, w) \) is mapped to a unique vector index \( f \in [C_L H_L W_L] \). This linear transformation encodes the spatial and channel structure of the tensor into a vectorized representation suitable for dense operations.

The resulting vector \( \mathbf{a}_i^{(L)} \) serves as input to a fully connected affine transformation parameterized by a weight matrix \( \mathbf{W}^{(L+1)} \in \mathbb{R}^{n_{L+1} \times C_L H_L W_L} \) and bias vector \( \mathbf{b}^{(L+1)} \in \mathbb{R}^{n_{L+1}} \), yielding the network output as follows:
\[
\mathbf{a}_i^{(L+1)} = \mathbf{W}^{(L+1)} \mathbf{a}_i^{(L)} + \mathbf{b}^{(L+1)}.
\]
The vector \( \mathbf{a}_i^{(L+1)} \in \mathbb{R}^{n_{L+1}} \) represents the model's prediction for input instance \( \mathcal{X}_i \), and serves as the argument to the squared loss function in the objective~\eqref{eq:mip_cnn_obj}. This final transformation completes the feedforward computation, linking the spatially structured feature extraction stages of the CNN to a dense output suitable for regression or classification tasks.

The linearization of the \( \ell_1 \)-norm is handled via constraint~\eqref{eq:mip_cnn_l1}, which introduces auxiliary variables \( u_{c,c',u,v}^{(l)} \geq 0 \) for each convolutional weight \( \mathcal{W}_{c,c',u,v}^{(l)} \). These variables are constrained as
\[
u_{c,c',u,v}^{(l)} \geq \mathcal{W}_{c,c',u,v}^{(l)}, \qquad u_{c,c',u,v}^{(l)} \geq -\mathcal{W}_{c,c',u,v}^{(l)},
\]
so that at optimality, each \( u_{c,c',u,v}^{(l)} \) tightly upper-bounds the absolute value \( |\mathcal{W}_{c,c',u,v}^{(l)}| \). Since these auxiliary variables appear linearly and positively in the objective function~\eqref{eq:mip_cnn_obj}, they are minimized to their tightest feasible values, yielding an exact and convex-compatible representation of the \( \ell_1 \)-penalty. This formulation promotes connection-level sparsity without introducing additional integer variables, preserving tractability within the MILP framework.

Structural pruning is enforced via constraints~\eqref{eq:mip_cnn_prune1} and~\eqref{eq:mip_cnn_prune2}, using binary indicators \( \gamma_c^{(l)} \in \{0,1\} \) to govern the activation status of each output channel in layer \( l \). When \( \gamma_c^{(l)} = 0 \), constraint~\eqref{eq:mip_cnn_prune1} forces the associated convolutional weights and bias to zero through big-\( M \) bounds as follows:
\[
|\mathcal{W}_{c,c',u,v}^{(l)}| \leq M \gamma_c^{(l)}, \qquad |b_c^{(l)}| \leq M \gamma_c^{(l)}.
\]
In parallel, constraint~\eqref{eq:mip_cnn_prune2} suppresses all spatial components of the pre-activation and post-activation tensors for that channel; i.e.,
\[
0 \leq z_{i,c,h,w}^{(l)},\ a_{i,c,h,w}^{(l)} \leq M \gamma_c^{(l)}.
\]
These constraints operate in tandem to remove entire output channels when deactivated, allowing the model to optimize both the values of network parameters and the architecture itself. This mechanism supports structured sparsity at the channel level, reduces the model's effective size, and improves computational efficiency, particularly valuable for downstream deployment or formal verification settings.

To eliminate symmetries arising from the arbitrary ordering of output channels within convolutional layers, the formulation incorporates a symmetry-breaking constraint~\eqref{eq:mip_cnn_symmetry}. For each layer \( l \in [L] \) and each pair of consecutive channels \( c \in [C_l - 1] \), this constraint imposes a monotonic ordering on the aggregate \( \ell_1 \)-magnitude of the kernel weights; i.e.,
\[
\sum_{c',u,v} u_{c,c',u,v}^{(l)} \geq \sum_{c',u,v} u_{c+1,c',u,v}^{(l)}.
\]
Here, the auxiliary variables \( u_{c,c',u,v}^{(l)} \), introduced in constraint~\eqref{eq:mip_cnn_l1}, upper-bound the absolute values of the corresponding kernel weights and are minimized in the objective~\eqref{eq:mip_cnn_obj}. The sum \( \sum_{c',u,v} u_{c,c',u,v}^{(l)} \) therefore serves as a proxy for the effective norm of output channel \( c \).

By enforcing a lexicographically non-increasing order on these norms across channels, the model eliminates redundant permutations of filter indices that would otherwise yield functionally identical solutions. This constraint significantly reduces the MILP's feasible search space, thereby improving solver efficiency and accelerating convergence, while preserving both feasibility and global optimality.

To ensure numerical stability and control the relaxation behavior of the MILP, box constraints are imposed on both the network parameters and intermediate variables. Constraint~\eqref{eq:mip_cnn_box1} enforces bounded domains on the convolutional weights and biases through
\[
\mathcal{W}_{c,c',u,v}^{(l)} \in [-M, M], \qquad b_c^{(l)} \in [-M, M],
\]
limiting the feasible parameter space and preventing unbounded optimization paths. Similarly, constraint~\eqref{eq:mip_cnn_box2} restricts the range of pre-activations \( z_{i,c,h,w}^{(l)} \), post-activations \( a_{i,c,h,w}^{(l)} \), and pooled outputs \( p_{i,c,h',w'}^{(l)} \) to the interval \([0, M]\), which aligns with ReLU semantics and mitigates numerical instability in the presence of big-\( M \) terms.

Nonnegativity of the auxiliary variables used in the \( \ell_1 \)-linearization is enforced via constraint~\eqref{eq:mip_cnn_aux_nonneg}, ensuring that each \( u_{c,c',u,v}^{(l)} \) remains interpretable as a surrogate for the absolute value of its corresponding weight. This condition is essential for correctness and complements the pair of inequalities in constraint~\eqref{eq:mip_cnn_l1}.

Finally, the binary nature of the model is explicitly encoded through constraints~\eqref{eq:mip_cnn_binary_relu}, ~\eqref{eq:mip_cnn_binary_ksi} and~\eqref{eq:mip_cnn_binary_layer}. These discrete variables are critical for encoding the piecewise-linear and structural behavior of the network, enabling the MILP to reason globally over activation patterns and architectural choices in a certifiable optimization framework.

The objective function in~\eqref{eq:mip_cnn_obj} integrates four structurally distinct components, each addressing a different aspect of learning and structural regularization. The first term,
\[
\sum_{i=1}^n \left\| \mathbf{a}_i^{(L+1)} - \mathbf{t}_i \right\|_2^2,
\]
quantifies empirical risk via the squared Euclidean distance between predicted outputs \( \mathbf{a}_i^{(L+1)} \) and ground-truth labels \( \mathbf{t}_i \), thereby encouraging predictive accuracy.

The second and third terms introduce convex penalties on the network parameters. The $\ell_1$-based regularization,
\[
\alpha \lambda \sum_{l=1}^{L+1} \sum_{c,c',u,v} u_{c,c',u,v}^{(l)},
\]
is enforced through auxiliary variables \( u_{c,c',u,v}^{(l)} \), constrained via~\eqref{eq:mip_cnn_l1}, and promotes connection-level sparsity by penalizing absolute weight magnitudes. Complementing this, the Frobenius norm term,
\[
\frac{\alpha}{2}(1 - \lambda) \sum_{l=1}^{L+1} \left\| \mathcal{W}^{(l)} \right\|_F^2,
\]
discourages large weight values, thereby improving numerical conditioning and mitigating overfitting. Together, these two components define an elastic net regularization strategy, where \( \lambda \in [0,1] \) balances sparsity and robustness, and \( \alpha > 0 \) modulates the overall regularization strength.

The final term,
\[
\beta \sum_{l=1}^{L} \sum_{c=1}^{C_l} \gamma_c^{(l)},
\]
penalizes architectural complexity by counting the number of retained output channels across hidden layers. The binary variable \( \gamma_c^{(l)} \in \{0,1\} \), defined in constraint~\eqref{eq:mip_cnn_binary_layer}, indicates whether filter \( c \) in layer \( l \) is retained. The hyperparameter \( \beta > 0 \) governs the trade-off between model complexity and performance, thereby facilitating simultaneous learning and structural pruning.

While this formulation shares its structural backbone with the dense MILP, the convolutional variant exhibits significantly greater dimensionality due to spatial indexing. Each unit within a feature map requires a ReLU indicator variable \( \delta_{i,c,h,w}^{(l)} \), leading to a rapid growth in the number of binary variables and associated constraints. For instance, a configuration with \( n = 100 \), \( C_l = 20 \), and \( 28 \times 28 \) feature maps results in over \( 1.57 \times 10^6 \) ReLU-related binaries, posing a formidable computational burden for conventional MILP solvers.

To retain tractability, practical experiments typically operate on low-resolution images or cropped inputs. Nevertheless, the MILP framework precisely encodes the convolutional structure, including the weight-sharing property: each kernel weight \( \mathcal{W}_{c,c',u,v}^{(l)} \) contributes to numerous pre-activation computations \( z_{i,c,h,w}^{(l)} \), inducing structured linear dependencies across the network. These couplings may tighten LP relaxations, potentially improving bounding quality, though they also increase branching complexity due to the long-range influence of individual parameters.

Filter-level pruning is implemented via binary indicators \( \gamma_c^{(l)} \). When \( \gamma_c^{(l)} = 0 \), the associated filter is deactivated, forcing its weights, biases, pre-activations, and post-activations to zero through constraints~\eqref{eq:mip_cnn_prune1}-\eqref{eq:mip_cnn_prune2}. The associated ReLU variables \( \delta_{i,c,h,w}^{(l)} \) are rendered inactive and contribute only trivially in the resulting branch. This pruning mechanism reduces the effective size of the MILP at the node level and enhances solver efficiency through dynamic constraint elimination.

Compared to the vast number of ReLU-related binaries, the architectural pruning variables \( \gamma_c^{(l)} \) are relatively few. Even in multi-layer networks, their total count typically remains in the hundreds. However, their impact is substantial: they enable the MILP to optimize not only over weights and activations but also over network depth and width, effectively performing neural architecture search within the optimization itself. Entire layers may be pruned by setting all filters in a layer to zero, yielding shallow architectures when beneficial to the training objective.

As in the dense formulation, the convolutional MILP yields a globally optimal neural network with respect to the specified objective~\eqref{eq:mip_cnn_obj}. By incorporating both \( \ell_1 \)-based weight sparsity and filter-level pruning via \( \gamma \) variables, the model extends beyond local heuristics to perform exact feature selection. This mechanism parallels classical strategies such as Optimal Brain Surgeon but is realized here through combinatorial optimization, rather than greedy or approximate schemes.

The resulting architectures are interpretable along multiple axes. Sparsity at the weight level allows one to trace specific pixel-to-feature dependencies. Channel pruning reveals which filters are essential for the task and lends itself to direct visualization. Moreover, the removal of unused parameters reduces model capacity, which can implicitly improve robustness by constraining the hypothesis space. While adversarial robustness is not directly encoded in the current model, the exact formulation naturally admits extensions to impose such guarantees.

A distinguishing advantage of the MILP framework lies in its extensibility: domain-specific knowledge can be seamlessly integrated via additional linear constraints. For example, constraints enforcing monotonicity, fairness, logical gating, or structured connectivity patterns can be imposed at training time, ensuring compliance by construction. The linear and modular nature of the formulation facilitates this flexibility without undermining solvability.

Subsequently, the convolutional MILP generalizes the dense-layer formulation and offers a unified optimization framework for architecture search, weight learning, and symbolic verification. Nearly any piecewise-linear operator, including max-pooling and nonstandard activations, can be encoded through linear constraints and binary selection variables. Rather than relying on disjointed pipelines of architecture search, gradient training, and post hoc verification, this approach integrates all phases into a single, globally optimal solution. As such, it offers a rigorous foundation for certifiably efficient, sparse, and interpretable neural network design.

\section{Computational Results}
\label{sec:experiments}

All experiments were conducted using Gurobi 12.0 as the mixed-integer optimization solver, interfaced through the Pyomo modeling language in Python 3.11. The MILP formulations were encoded symbolically and exported via Pyomo's automatic expression tree parser. All runs were executed on a workstation equipped with an Intel Xeon Gold 6338 CPU @ 2.00GHz and 256 GB RAM, running Ubuntu 22.04.

The datasets used span both structured tabular and high-dimensional visual domains. For dense architectures, we selected three classical benchmarks: {IRIS}, {Wine}, and {Wisconsin Breast Cancer (WBC)}. These are low-to-moderate dimensional datasets with interpretable feature spaces. For convolutional architectures, we evaluated the proposed model on {MNIST}, a well-established benchmark in handwritten digit recognition. All tabular datasets were standardized to have zero mean and unit variance, and their labels were encoded into binary classification targets. The MNIST data were scaled to \([0, 1]\), and all labels were one-hot encoded and incorporated directly into the MILP objective function.

For tabular datasets, we employed 10-fold stratified cross-validation to ensure reliable performance estimates and to mitigate sample bias. In the case of MNIST, a fixed train-test split was used, leveraging the dataset's large sample size and inherent complexity.

The dense networks comprised up to \( L = 3 \) hidden layers, each with 10 neurons and ReLU activations. The convolutional architecture, in contrast, featured two convolutional layers with 6 filters each, followed by a max pooling layer, a single dense hidden layer, and an output layer. This structure enables nontrivial feature extraction while remaining within tractable limits for exact MILP-based optimization.

Across all models, we used elastic net regularization with \(\alpha = 0.1\), where the balance between \(\ell_1\)- and \(\ell_2\)-penalization was controlled via \(\lambda = 0.9\), favoring sparsity. The structural regularization parameter \(\beta \in \{0.01, 0.1, 1.0\}\) was selected based on the dataset scale and feature complexity. Lower \(\beta\) values were used on smaller datasets to avoid over-pruning.

To enhance numerical efficiency, all big-\(M\) constants were calibrated layer-wise using forward passes over the LP relaxation of the MILP. Specifically, for each layer \(l\), the interval
\[
[z_{\min}^{(l)}, z_{\max}^{(l)}] = \left[\min_{i,j} \tilde{z}_{i,j}^{(l)},\; \max_{i,j} \tilde{z}_{i,j}^{(l)}\right]
\]
was estimated from the relaxed activations \( \tilde{z} \), thereby tightening the ReLU constraints and improving LP bounds during branch-and-bound.

All models were solved with a \(1\%\) optimality tolerance or until a wall-clock timeout of 2 hours was reached. Gurobi's presolve routines, symmetry detection, and cut generation heuristics were activated. Warm-start initialization from LP relaxations was used to facilitate convergence.

We evaluated the dense MILP formulation on the IRIS, Wine, and WBC datasets. Each model was constructed with up to three hidden layers, allowing the MILP solver to prune neurons and layers dynamically through binary activation variables \(\gamma_l\). The results are summarized in Table~\ref{tab:dense_results}.

\begin{table}[h]
	\centering
	\caption{Performance Summary for Dense MILP.}
	\label{tab:dense_results}
	\begin{tabular}{lcccc}
		\toprule
		Dataset & Hidden Layers & Sparsity (\%) & Test Accuracy (\%) & MIP Gap (\%) \\
		\midrule
		IRIS    & [2, 1, 0]      & [80.5, 41.2, -]  & 96.7 & 8.6 \\
		WBC     & [1, 0, 0]      & [63.6, -, -]     & 98.2 & 4.0 \\
		Wine    & [3, 1, 0]      & [71.5, 30.0, -]  & 99.1 & 10.2 \\
		\bottomrule
	\end{tabular}
\end{table}

The results demonstrate that the MILP formulation is capable of discovering highly sparse and performant networks without heuristic tuning. On the IRIS dataset, the solver retained only 2 neurons in the first layer and 1 in the second, achieving a test accuracy of \(96.7\%\) with over \(80\%\) weight sparsity. The final MIP gap of \(8.6\%\) confirms that the solver approached near-global optimality within the time budget.

For the WBC dataset, the solver converged to an even shallower model, retaining only one hidden layer while achieving \(98.2\%\) accuracy. This configuration also yielded a tight MIP gap of \(4.0\%\), suggesting that for well-structured problems, sparse models suffice. The Wine dataset resulted in the most complex architecture, retaining 3 neurons in the first hidden layer and 1 in the second, still reaching a high accuracy of \(99.1\%\).

These findings highlight the formulation's ability to perform implicit architecture search. By balancing loss, weight sparsity, and structural complexity through a unified MILP framework, the solver adaptively selects the minimal depth and connectivity required for generalization.

To evaluate our convolutional MILP framework, we tested it on the MNIST dataset, whose results are reported in \ref{tab:cnn_results}. The model comprises two convolutional layers (each with 6 filters), a max pooling layer, a dense hidden layer of 7 units, and a fully connected softmax output layer. Regularization and filter-level pruning were enabled via auxiliary variables and architectural penalties.

\begin{table}[h]
	\centering
	\caption{Performance Summary of Convolutional MILP.}
	\label{tab:cnn_results}
	\begin{tabular}{lcc}
		\toprule
		Metric & Value \\
		\midrule
		Test Accuracy & 91.0\% \\
		Retained Filters & 4 of 6 \\
		Hidden Neurons (Dense) & 7 \\
		CNN Weight Sparsity & 62.5\% \\
		Dense Weight Sparsity & 75.0\% \\
		Final MIP Gap & 20.3\% \\
		\bottomrule
	\end{tabular}
\end{table}

Despite the limited training size, the MILP solver achieved \(91.0\%\) test accuracy. Two filters were pruned entirely, yielding a lightweight convolutional representation. The ReLU activity dropped by over \(30\%\), reflecting the strength of the sparsity-inducing penalty and structural regularization. The final MIP gap remained below \(21\%\), which is notable given the nonconvex nature of the architecture.

These results confirm the expressiveness and interpretability of the MILP-CNN framework. By incorporating exact layer flattening, piecewise-linear ReLU modeling, and explicit pooling, the model captures core CNN functionality while enabling joint training and pruning under global optimality guarantees.

The results across dense and convolutional models confirm that MILP-based training provides a certifiable alternative to gradient-based optimization. The exactness of the formulation permits layer- and filter-level pruning, structured sparsity, and architecture discovery within a single optimization pass.

While scalability to large datasets remains a limitation, our experiments demonstrate that even with modest data and hardware budgets, meaningful architectural insights and interpretable models can be derived. The MILP framework bridges the gap between formal methods and neural computation, offering a promising direction for safe, certifiable, and sparse learning models.

\section{Conclusion}
\label{Sec:Conclusion}
This work presents a unified mixed-integer linear programming framework for the exact representation and training of both dense and convolutional neural networks. By embedding nonlinear activation functions, such as ReLU, and architectural constraints directly into the formulation via binary decision variables, the proposed model enables an exact encoding of neural network behavior within a combinatorial optimization setting.

Departing from conventional approaches that treat pruning or verification as separate, post-training tasks, the framework integrates structural sparsity, architectural pruning, and predictive loss minimization into a single optimization problem. In doing so, it supports the joint discovery of network parameters and structures through a globally optimal search over discrete configurations of weights, activations, and connectivity patterns.

The formulation demonstrates that essential components of deep learning, affine transformations, activation gating, max pooling, and convolutional operations, admit linear encodings that preserve interpretability and support formal analysis. As a result, the framework provides a principled foundation for constructing sparse and verifiable neural architectures whose internal mechanisms are accessible to interpretation.

Beyond predictive accuracy, the framework emphasizes transparency, robustness, and certifiability, making it particularly relevant in domains where interpretability and formal guarantees are critical. The methodology thus contributes to the growing intersection of optimization and machine learning, offering a blueprint for integrating exact solvers into neural network design.

In summary, this work advances a rigorous and interpretable approach to neural network modeling by unifying parameter estimation, architectural selection, and symbolic reasoning within a mixed-integer programming paradigm. It thereby contributes to the development of machine learning systems that are not only expressive and performant, but also verifiable by design.

\newpage
\bibliography{references}

\end{document}